# Combining Particle Swarm Optimizer with SQP Local Search for Constrained Optimization Problems

Pelley C., Innocente M., Sienz J. *

*Keywords: Particle swarm optimization (PSO), Sequential quadratic programming (SQP), Local search hybrid, GP-PSO*

The combining of a General-Purpose Particle Swarm Optimizer (GP-PSO) with Sequential Quadratic Programming (SQP) algorithm for constrained optimization problems has been shown to be highly beneficial to the refinement, and in some cases, the success of finding a global optimum solution. It is shown that the likely difference between leading algorithms are in their local search ability. A comparison with other leading optimizers on the tested benchmark suite, indicate the hybrid GP-PSO with implemented local search to compete along side other leading PSO algorithms.

## Nomenclature

| | | |
|---|---|---|
| PSO | = | Particle Swarm Optimizer |
| GP-PSO | = | General Purpose PSO |
| SQP | = | Sequential Quadratic Programming |
| GP-PSO-SQP | = | General-Purpose PSO with SQP |
| GA | = | Genetic Algorithm |
| EA | = | Evolutionary Algorithm |
| MA | = | Memetic Algorithm |
| SA | = | Simulated Annealing |
| SI | = | Swarm Intelligence |
| DMS-PSO | = | Dynamic Multi-Swarm PSO |
| PESO+ | = | Particle Evolutionary Swarm Optimization Plus |
| FE | = | Function Evaluations |

*School of Engineering, Swansea University, Singleton Park, SA2 8PP, UK. Pelley C. Email: 364915@swansea.ac.uk; Innocente M. Email: m.s.innocente.356695@swansea.ac.uk; Sienz J. Email: J.Sienz@swansea.ac.uk

# 1 Introduction

The main motivation and inspiration of this investigation is regarding the 'Law of Sufficiency' as described by Kennedy et al.[1] (p175)"If a solution is good enough, and it is fast enough, and is cheap enough, then it is sufficient". Following this line of thought, comparisons between General Purpose PSO and GP-PSO-SQP are made together with two of the leading algorithms.

## 1.1 Particle Swarm Optimizer

Research into the field of Swarm Intelligence (SI) in the last two decades has resulted in many advantages being exploited for the purpose of optimization. SI fits into the category of modern heuristics, defined here as an algorithm that intends to find a solution to a problem with suitable computational time without guarantee of optimality. The original formalization of the PSO paradigm and its place amongst other paradigms was first described by Kennedy et al.[2]. Kennedy et al. describes PSO as fitting into these categories and to have roots in Evolutionary Algorithms (EA) and Genetic Algorithms (GA). Stochastic processes giving it similarity to the former and an ability to follow a local and neighbourhood best being similar to the crossover operator in the later. The advantage of modern heuristics over that of traditional methods is that they are not problem specific.

A general description of the original PSO by Kennedy et al.[2][3] is a randomly initialized swarm within feasible space with randomly initialised velocities. The velocity of each of the $n-dimensional$ particles is accelerated towards its own personal best position and towards the best of its local neighbourhood with stochastic weighting between the former and the later.

Let $v_{ij}^{(t)}$ be the velocity at the current time-step and $x_{ij}^{(t)}$ be the current solution coordinate, then we have the basic underlying equations driving the GP-PSO:

$$v_{ij}^{(t)} = w \cdot v_{ij}^{(t-1)} + iw \cdot U_{(0,1)} \cdot \left(pbest_{ij}^{(t-1)} - x_{ij}^{(t-1)}\right) + sw \cdot U_{(0,1)} \cdot \left(lbest_{ij}^{(t-1)} - x_{ij}^{(t-1)}\right) \quad (1)$$

$$x_{ij}^{(t)} = x_{ij}^{(t-1)} + v_{ij}^{(t)} \quad (2)$$

In Eqn.1, $U_{(0,1)}$ is a random number from a uniform random distribution in the interval [0,1], sampled anew for each time it is called; $w$, $iw$ and $sw$ are the inertia, individuality and social weights respectively; **pbest**$_i$ and **lbest**$_i$ are the solution coordinate of the i'th particles best ever position and the solution coordinate of the best of the i'th particle's neighbourhood respectively.

A number of the coefficients in PSO, though small in number, are highly problem dependant and as such, the GP-PSO developed by Innocente et al.[4] can be characterised by the features developed to overcome this inherent problem and can produce high quality solutions to trigger a local search.

A brief overview of features of the GP-PSO designed by Innocente et al.[4] is described as follows; The set-up consists of three swarms of different behaviour (different coefficients), complementing each other on the others weaknesses whether exploitative or explorative. The neighbourhood is of a 'forward topology' as proposed by Innocente et al.[4], similar to that of the ring[1] topology only that interconnections are not bidirectional in the former. The intention is to slow clustering of the swarm and increase exploration of the search space. The size of these neighbourhoods is dynamic in that they change linearly with time so that the search begins highly local (extending the search of the space and delaying clustering), then becoming a global search toward the end with total cooperation between particles.

Constraints are handled by the preserving feasibility method with priority rules and pseudo-adaptive relaxation of the tolerances for both equality and inequality constraint violations, as proposed in [4].

### 1.2 Sequential Quadratic Programming

Combining the PSO algorithm and SQP gradient by providing the latter with 'good' initial solutions, enables the feature of guaranteed local optima convergence which the PSO alone does not have. SQP is known to be one of the most successful methods in non-linear constrained optimization, as discussed by a number of authors in the literature, for example, by Victoire et al., who investigated a hybrid PSO-SQP algorithm for an economic dispatch problem [5]. It seems a logical approach for refined local search hybridization with heuristic algorithms on constrained optimization problems. SQP is a quasi-newton method utilising second-order information about the problem to efficiently and accurately converge to a solution. The method considered is an active-set algorithm for dealing with constrained optimization problems by finding a solution to the Karush-Kuhn-Tucker (KKT) conditions, which are analogous to finding a point at which the gradient is zero, only considering constraints as well. At each iteration, an approximation of the Hessian of the Lagrangian is made using the Broyden-Fletcher-Goldfarb-Shanno (BFGS) method. This is then used to formulate a general quadratic problem (QP) whose solution is used to determine a direction, in which a line search is performed. This is formulized by bound constraints expressed as inequality constraints and non-linear constraints linearized. This procedure is then repeated until some termination criteria have been met.

## 2 Related Work

Memetic algorithms (MA) were first proposed by Moscato in 1989 (quoted by Petalas et al. in [6]), being inspired by the notion of Memes, proposed by Dawkins in 1976 as a unit of cultural evolution (quoted in [6]). Moscato implemented a hybrid population-based GA with Simulated Annealing (SA) as a local search refinement to tackle combinatorial problems including the travelling salesman problem. The method gained wide acceptance due to its ability to solve difficult[2] problems. A comparative study was made by Petalas et al.[6] on MAs, concluding them to be highly superior in effectiveness compared to the stand-alone global algorithm.

---

[1]Ring topology is where each particle shares information with only two other particles in the swarm.

[2]NP-hard problems are a particular class of combinatorial problems that no polynomial time algorithm exists and so computation times may tend to exponential computation time in the worst possible cases as described by Christian B [7].

Implementations of hybrid based methods are far from uncommon amongst the optimization community. Examples include the training of artificial neural networks for function approximation[8] or the hybrid SQP-PSO designed by Victoire et al.[5] for the economic dispatch problem. The latter being of high interest with its implementation of SQP to the current best solution found in the swarm (**gbest**), triggered each time the solution has improved. It was argued by Victoire et al.[5] that early on in the PSO search, particles are statistically likely to be of proximity to the global best but then move away from these areas. For this reason the local search method was implemented.

One of the most successful algorithms developed for a general purpose optimizer is by Liang et al.[9], that combines and in fact couples the SQP method with their Dynamic Multi-swarm Optimizer (DMS-PSO). Their method describes subpopulations solving their own objectives, being assigned adaptively, and their assignment being periodically changed according to difficulty. For this reason, the number of subpopulations is not necessarily equal to the number of objectives or constraints. The SQP method is coupled by being called at every '$n$' number of generations, where the positions of the individual best experiences of five randomly chosen particles comprise the seeds for the local searches. The random choosing of these five particles means that no preference is made to one over the other, which is consistent with the fact that particles distances from an unknown global optimum give rise to indistinguishability between one 'good'[3] particle and another. After a certain percentage of function evaluations are met, every '$n$' generations, a local search is triggered using the **gbest** solution only (thus ensuring refinement of the final solution).

## 3 Experimental

A set of problems are used for the purpose of testing the hybrid algorithm's effectiveness to solve real-world optimization problems. These test suites include the 13 problems taken from Pulido et al.[10] (referred hereafter as problems g01-g13) with some added features and performance measures in [11] (referred hereafter as problems g14-g24). Measures determining clustering of the swarm considered are those formulated by Innocente et al.[12].

With respect to tolerance, to remain consistent with problem definitions for CEC06[11], equality constraints are relaxed by formulating them as inequalities as shown below:

$$|h_j(\mathbf{x})| - \epsilon \leq 0 \qquad (3)$$

Where $\epsilon = 10^{-4}$

Inequalities are then defined as;

$$g_j(\mathbf{x}) \leq 0 \qquad (4)$$

---

[3]'good' particle refers to the quality of a particles solution, considering both its conflict and constraint.

It should be noted that all constraints once formulated to inequalities have zero tolerance for the GP-PSO, but within the SQP, due to round-off errors, constraints are considered not violated if $\leq 10^{-12}$.

Tolerances for the SQP are chosen based on a sample of problems of the test suite, where the ability of the SQP to converge and terminate at a solution of the desired level of accuracy, result in the following tolerances set: $Tolx = 10^{-12}$ (tolerance on the solution coordinates); $Tolcon = 10^{-14}$ (tolerance on the constraints); $Tolfun = 10^{-14}$ (tolerance on the conflict function).

The SQP local search is applied to the final results of the GP-PSO on problems g01-g24 and a determination is made to whether a local search is helpful or not. Secondly, a local search triggering is made on each iteration of the GP-PSO on problems g01-g13, to determine the points at which the SQP local search becomes successful. With the results obtained from the above, considerations are made to difficulties met on the 13 problems (where the local search fails and where might it be sensitive) with particular attnetion made to low dimensional problems. Finally, a comparison between the results obtained for the entire benchmark and the two most successful algorithms from the CEC06[11] test suite are made to determine the validity of a local search hybrid together with the possible computational expense of the algorithms.

## 4  Results and Discussion

Comparison between the path of the SQP local search at various points in the solution space is made with that of the GP-PSO on low dimensional problems including g06, g08 and g11. All three functions offer significant insight into the advantages of local search implementation to the heuristic global optimizing PSO search, together with the drawbacks of traditional methods alone.

In problem g06, the path of the SQP with various initial solutions is shown in Fig.[1,2]. The chosen starting points are based on varied feasible and infeasible points in the search-space.

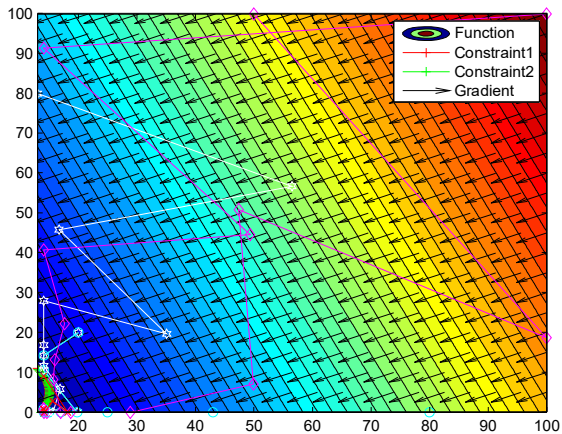 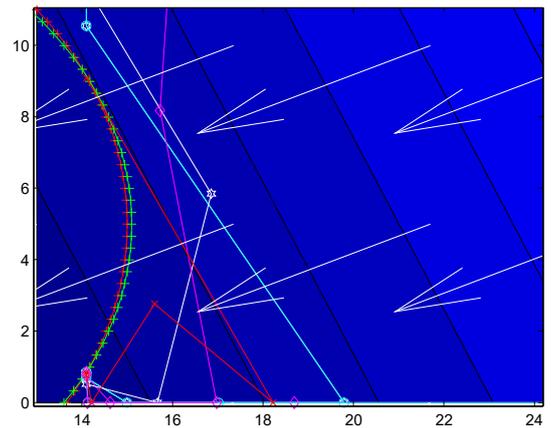

Figure 1: g06 GP-PSO path (full search-space)    Figure 2: g06 GP-PSO path (region of interest)

The results indicate that the SQP tends to directions unexpected depending on the problem features. In this case, the edge of the search-space is seen as an attractive area since it offers an improved conflict and a lower maximum constraint violation. It takes a number of iterations before this local optimal region is escaped, where the search for a feasible region of the search-space and a degradation of the conflict is observed to reach this feasible region. Since this approach takes the SQP path through a local optimal region of the search-space, it is apparent that the initial starting solution to the local search will determine the final converged solution in more complicated problems (suggesting that an erratic success/failure may occur in some cases).

The path of the GP-PSO as shown in Fig.3 indicates a similarity with the SQP path, with the best solution history of the swarm following the edge of the search-space for a time before convergence to the global optimum occurs. This similarity is however expected, since the PSO is statistically likely to follow the path of the gradient to optimality. It is noticed that since the PSO relies on its diversity and statistical likelihood in finding a better solution other than the boundary edge, that the SQP offers to be a more efficient method in dealing with this function. The local search method is clearly a more efficient method in dealing with this function.

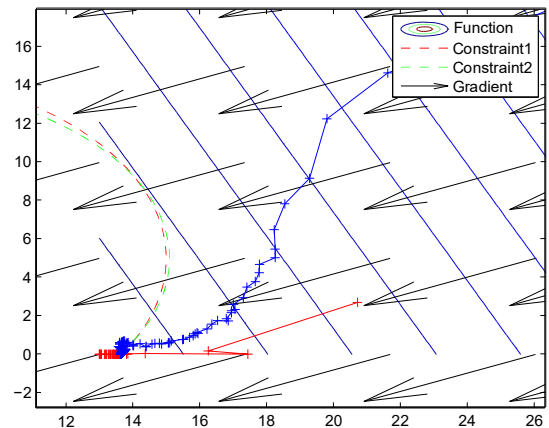

Figure 3: g06 GP-PSO path (region of interest), Red curve showing the global best history and Blue curve the centre of gravity history.

Problem g08 offers to be a highly interesting function, in that it is clearly multimodal, and as such a local search becomes easily trapped. This is clearly shown in Fig.4. There are a number of suboptimal regions within the feasible search-space and the conflict is rather flat (difference between the extremities within the region is small). The SQP can hop over the global best in the pursuit of satisfying the constraints. It is also apparent that the SQP may be led to a solution that may cause an error with a divide by zero even though the formulization of this problem allows solution coordinates of (0,0). Initial starting points near the global optimum still do not guarantee convergence on the global optimum since following the curvature of the constraint function leads it away, indicating a lack of knowledge of the function due to a highly localised approximation of the conflict and constraint function in the lagrangian. There are at least 6 suboptimal regions within the feasible search-space that the SQP may become trapped. The SQP also often finds itself stuck in the suboptimal regions at the left boundary of the search-space.

The path of the GP-PSO as shown in Fig.5, demonstrates its similarity to the SQP path in that it is pulled toward the edge of the search-space. Since diversity is kept, the best solution found by the swarm is quickly shifted to a region of feasibility by other members of the swarm searching the solution space. The path of the PSO demonstrated its success in finding a global optimum in a search-space where the local SQP optimizer is prone to fail.

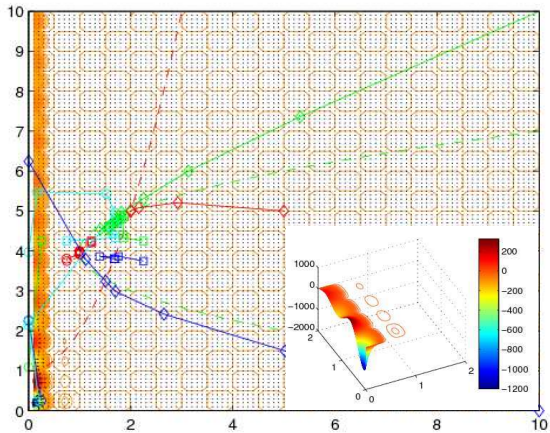
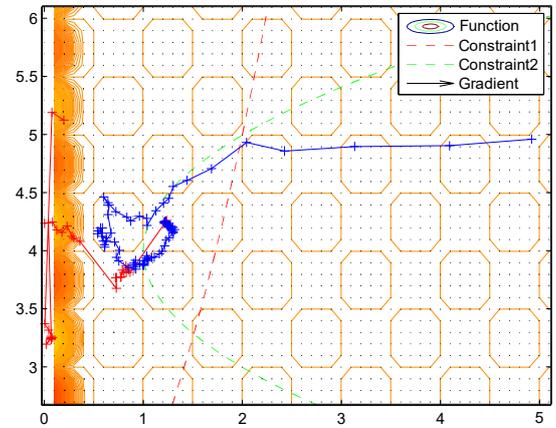

Figure 4: g08 SQP path (full search-space), with a 3D plot of small region of the search-space to better understand the problem features.

Figure 5: g08 GP-PSO path (region of interest), Red curve showing the global best history and Blue curve the centre of gravity history.

In problem g11, it is clearly shown in Fig.[6,7] that a local search on this simple function does not guarantee global optima convergence. Depending on the initial starting solution, it is possible for the SQP to become trapped at the local optimum at the centre of the search-space. It is however unlikely for this to happen unless the starting solution is nearly on top of it. The way in which the SQP escapes this region if near it, is to improve its conflict at the cost of increasing its constraint violation and then to do the vice versa, and so, zigzag its way to optimality. Due to the central region of sub-optimality it takes a number of iterations to escape this region with respect to other areas of the search-space. The formulation of the equalities and inequalities is beneficial in the case of this function, since it allows the solution to deteriorate a little, however equality relaxation may be harmful as well as beneficial. It may open a suboptimal region for which the SQP may become trapped or on the other hand, open regions normally separated by a worsening solution thus making them accessible.

It is also useful to understand how the SQP might deal with multimodal problems (where in this case, this vertically symmetric function offers two equally 'good' global optima solutions). It is noticed that depending on the initial solution provided to the SQP, the chosen global optimum is determined (since the SQP method is a deterministic algorithm). For initial solutions at the left or right border edge of the solution space it is noticed that the initial path of the SQP take it toward the corners where the constraints are satisfied. This convex function is highly appropriate in demonstrating the success of the local search SQP to find the global optimum.

The path of the GP-PSO as shown in Fig.8 demonstrates the effect of two equally good attractors on the swarm behaviour. Unlike the path of the SQP, the swarm's best solution fluctuates between the two global optima since the diversity of the swarm causes better solutions to be found on either side of this vertically symmetric function. The centre of gravity solution of the swarm begins at the centre with the swarm having been randomly initialized, and then it tends to a point which is vertically cantered and horizontally in line with the two solutions. This is due to the two attractors being equally 'good'. After a time, a more refined solution on either side will become less and less frequent, at which point the stochastic choice of the global optima will cause the centre of gravity to drift toward it. It is again apparent that the SQP deals with this function much more efficiently to the GP-PSO.

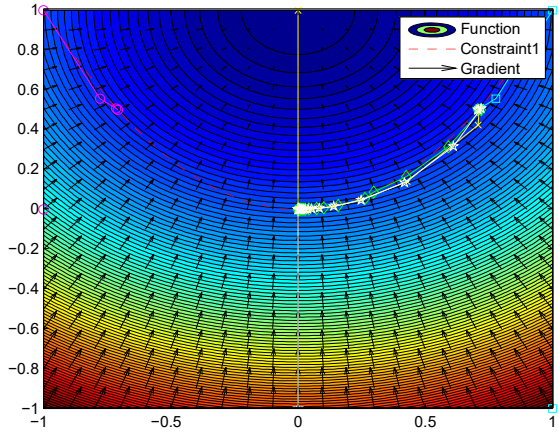 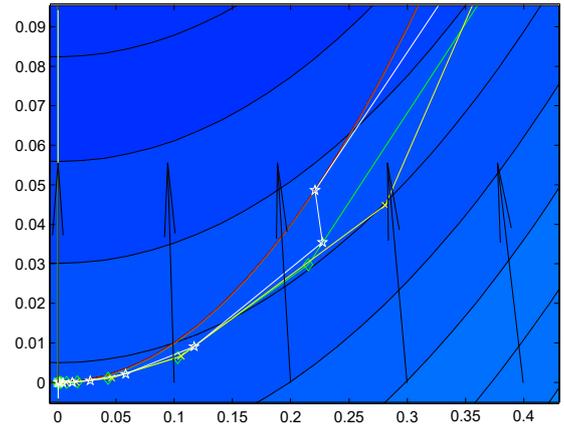

Figure 6: g11 problem for SQP applied at various starting positions (full search-space)

Figure 7: g11 problem for SQP applied at various starting positions (region of interest)

As shown by the low dimensional problem investigation, features of problems suggest that a MA to be highly beneficial, however that both complement each other on each others weaknesses. The PSO relies on its statistical likelihood to find a 'good' solution while the local search is shown to be reliant on the initial solution it is provided with.

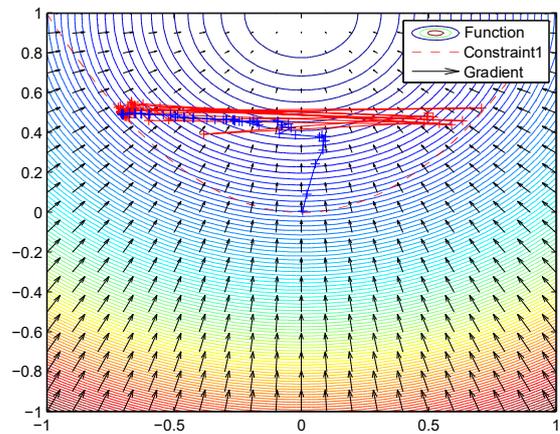

Figure 8: g11 GP-PSO path (region of interest), Red curve showing the global best history and Blue curve the centre of gravity history.

Combining SQP with the GP-PSO, indicates much improvement of its solutions over the standalone GP-PSO algorithm. GP-PSO is found to fail on problems g05, g07, g09, g10. The success of the SQP on all of these problems when applied to the GP-PSO final iteration is then a significant result. An investigation into the point at which the SQP becomes successful is made by applying the local search at each iteration.

Problems g05, g10 and g13 have erratic success rates when a local search is applied due to their sensitivity to scaling. Problems g03, g06, g09 and g11 however are successful from the very first iteration.

The results of SQP implementation to the final solution of the GP-PSO is shown in Table.3 and the success and feasibility rate shown in comparison to that of the stand-alone GP-PSO together with two of the leading optimizers shown in Table.1. The results indicate that the SQP gives not only an improved refined solution but also in some problems leads to a solution where none are found by the stand-alone GP-PSO.

| PROBLEM | GP-PSO SUCCESS | FEASIBLE | GP-PSO-SQP SUCCESS | FEASIBLE | PESO+ SUCCESS | FEASIBLE | DMS-PSO SUCCESS | FEASIBLE |
|---|---|---|---|---|---|---|---|---|
| g01 | 100% | 100% | 100% | 100% | 100% | 100% | 100% | 100% |
| g02 | 70% | 100% | 70% | 100% | 56% | 100% | 84% | 100% |
| g03 | 70% | 100% | 100% | 100% | 100% | 100% | 100% | 100% |
| g04 | 100% | 100% | 100% | 100% | 100% | 100% | 100% | 100% |
| g05 | 0% | 100% | 100% | 100% | 100% | 100% | 100% | 100% |
| g06 | 100% | 100% | 100% | 100% | 100% | 100% | 100% | 100% |
| g07 | 0% | 100% | 100% | 100% | 96% | 100% | 100% | 100% |
| g08 | 100% | 100% | 100% | 100% | 100% | 100% | 100% | 100% |
| g09 | 0% | 100% | 100% | 100% | 100% | 100% | 100% | 100% |
| g10 | 0% | 100% | 70% | 70% | 16% | 100% | 100% | 100% |
| g11 | 100% | 100% | 100% | 100% | 100% | 100% | 100% | 100% |
| g12 | 100% | 100% | 100% | 100% | 100% | 100% | 100% | 100% |
| g13 | 90% | 100% | 100% | 100% | 100% | 100% | 100% | 100% |
| g14 | 0% | 100% | 100% | 100% | 0% | 100% | 100% | 100% |
| g15 | 90% | 100% | 100% | 100% | 100% | 100% | 100% | 100% |
| g16 | 100% | 100% | 100% | 100% | 100% | 100% | 100% | 100% |
| g17 | 100% | 100% | 90% | 90% | 0% | 100% | 0% | 100% |
| g18 | 20% | 100% | 100% | 100% | 92% | 100% | 100% | 100% |
| g19 | 0% | 100% | 100% | 100% | 0% | 100% | 100% | 100% |
| g20 | NA | 0% | NA | 0% | NA | 0% | NA | 0% |
| g21 | 0% | 0% | 0% | 0% | 0% | 100% | 100% | 100% |
| g22 | 0% | 0% | 0% | 0% | 0% | 0% | 0% | 0 |
| g23 | 0% | 0% | 100% | 100% | 0% | 96% | 100% | 100% |
| g24 | 100% | 100% | 100% | 100% | 100% | 100% | 100% | 100% |

Table 1: Final success rates and feasibility rates using the end values (corresponding to 10,000 iterations in the case of the GP-PSO and GP-PSO-SQP.

Problems g05, g07, g09, g10, g14, g19 and g23 are shown to improve considerably in their solution quality when the local search is applied. It is likely that the guaranteed local optima convergence, overcomes the inherent limitation of the stochastic swarm, where it is shown to refine or even find the solution (to within accuracy of results defined by CEC06). Since the swarm relies on its diversity, when it is lost (depending on the features of the problem), the swarms centre of gravity solution and the best solution may not ever converge to the global optimum as the swarm becomes as one particle and loses momentum. The local search implementation then allows the local convergegence within machine precision thus overcoming this limitation. The most notable problems however in its implementation are in problems g05, g07, g10 and g17, where scaling and overshooting (termination criteria) are found to be a major issue in the outcome of the converged solution. It is also noted that GP-PSO struggles considerably in obtaining feasible solutions to problems g21-g23, however the SQP method results in a 100% on problem g23. The lack a feasible solution in problems g21-22 results in total failure of the SQP to find a feasible solution either. PESO+ also fails on these two problems but DMS-PSO is successful on problem

g21.

With respect to the leading optimizers, the simplified GP-PSO indicates that it belongs alongside the two leading algorithms with its success on g17 which both PESO+ and DMS-PSO fail. Success of the SQP when supplied with initial 'good' solutions by the GP-PSO on problems g14, g19 and g23, indicate that the local search implementation to be key to the success and refinement of solutions for a general purpose optimizer since only the GP-PSO-SQP and the DMS-PSO meet success on these problems. The implementation of the SQP local optimizer, brings it in line with the two leading algorithms and a local search is likely to bridge the gap between the two leading optimizers. It is however indicated that since the GP-PSO fails to find a feasible solution on problem g21, it is speculated that perhaps recent developments implemented by Innocente et al.[4] such as the repair operator, may help improve the solutions provided to the SQP, and as such lead to at least feasible solutions on the search-space.

|     | GP-PSO  | GP-PSO_loc | SQP     | PESO+   | DMS-PSO |
|-----|---------|------------|---------|---------|---------|
| g01 | 5.5E+04 | 3.1E+04    | 9.2E+01 | 1.0E+05 | 3.3E+04 |
| g02 | 1.7E+05 | 1.1E+05    | 1.3E+03 | 2.3E+05 | 1.8E+05 |
| g03 | 3.2E+04 | 1.7E+03    | 1.3E+03 | 4.5E+05 | 2.6E+04 |
| g04 | 4.2E+04 | 4.1E+04    | 3.2E+01 | 8.0E+04 | 2.5E+04 |
| g05 | –       | 0.0E+00    | –       | 4.5E+05 | 2.9E+04 |
| g06 | 4.1E+04 | 0.0E+00    | 4.0E+01 | 5.7E+04 | 2.8E+04 |
| g07 | –       | 6.1E+04    | 5.5E+02 | 3.5E+05 | 2.7E+04 |
| g08 | 1.2E+04 | 1.1E+04    | 8.5E+01 | 6.1E+03 | 4.1E+03 |
| g09 | –       | 0.0E+00    | 2.8E+02 | 9.8E+04 | 2.9E+04 |
| g10 | –       | 7.2E+04    | 7.2E+02 | 4.5E+05 | 2.6E+04 |
| g11 | 1.0E+04 | 0.0E+00    | 4.0E+01 | 4.5E+05 | 1.5E+04 |
| g12 | 9.1E+03 | 6.1E+03    | 4.1E+01 | 8.1E+03 | 5.4E+03 |
| g13 | 4.7E+04 | 4.4E+03    | 1.6E+02 | 4.5E+05 | 4.1E+04 |
| g14 | –       | 5.2E+04    | 1.5E+03 | –       | 2.5E+04 |
| g15 | 3.8E+04 | 0.0E+00    | 8.2E+01 | 4.5E+05 | 2.9E+04 |
| g16 | 2.3E+04 | 2.5E+04    | 1.1E+02 | 4.9E+04 | 5.3E+04 |
| g17 | 7.4E+04 | 8.2E+04    | 1.5E+03 | –       | –       |
| g18 | 4.5E+04 | 2.0E+04    | 2.0E+02 | 2.1E+05 | 3.3E+04 |
| g19 | –       | 1.3E+04    | 4.2E+02 | –       | 2.2E+04 |
| g20 | NA      | NA         | NA      | NA      | NA      |
| g21 | –       | –          | –       | –       | 1.4E+05 |
| g22 | –       | –          | –       | –       | –       |
| g23 | –       | 3.7E+04    | 2.6E+02 | –       | 2.1E+05 |
| g24 | 1.2E+04 | 7.4E+00    | 2.9E+01 | 2.0E+04 | 1.9E+04 |

Table 2: Mean number of FES to achieve the fixed accuracy level $((f(\bar{x}) - f(\bar{x}^*) \leq 1e-4))$. GP-PSO_loc represents the mean FE at which the local search becomes successful. SQP is the mean number of FEs required at the point at which it becomes successful.

With respect to computational expense, Table.2 indicates the mean function evalutations (FE) on the bench mark problems, between the three algorithms (GP-PSO, DMS-PSO and PESO+) together with the FEs required from the point at which a local search becomes successful. When compared on problems successful between all three algorithms (GP-PSO with local search implementation included for completeness), the mean FEs put the DM-PSO in the lead with the GP-PSO closely behind (with the GP-PSO's success likely being due its not having to calculate the objective function outside the feasible space). PESO+ has indicated that it requires a great deal more FEs on average with respect to the other optimizers, however it is a considerably more robust optimizer than the standard GP-PSO with its success on a greater number of problems of the test suite. From the application of the SQP, it is clear that on average, the number of FEs required before error is attained is halved with respect to that required for the basic GP-PSO, and that the number of FEs by the local search itself is on average only 2% the total computational cost. A trigger is however required in order to switch between the global and local search.

| PROBLEM | OPTIMUM | | GP-PSO Conflict | Max Constraint | SQP SQPconf | SQPcons |
|---|---|---|---|---|---|---|
| g01 | -15.000000 | BEST | -15.000000 | 0.0E+00 | -15.000000 | 0.0E+00 |
|  |  | AVERAGE | -15.000000 | 0.0E+00 | -15.000000 | 0.0E+00 |
|  |  | STDEV | 5.8E-12 | 0.0E+00 | 2.9E-12 | 0.0E+00 |
| g02 | -0.803619 | BEST | -0.803616 | 0.0E+00 | -0.803619 | 3.1E-15 |
|  |  | AVERAGE | -0.800309 | 0.0E+00 | -0.800316 | 4.3E-15 |
|  |  | STDEV | 5.3E-03 | 0.0E+00 | 5.3E-03 | 2.2E-15 |
| g03 | -1.000500 | BEST | -1.000495 | -5.2E-06 | -1.000500 | 7.0E-16 |
|  |  | AVERAGE | -1.000102 | -5.4E-07 | -1.000500 | 5.2E-16 |
|  |  | STDEV | 9.9E-04 | 1.6E-06 | 2.7E-15 | 6.7E-16 |
| g04 | -30665.538672 | BEST | -30665.538672 | 0.0E+00 | -30665.538672 | 0.0E+00 |
|  |  | AVERAGE | -30665.538672 | 0.0E+00 | -30665.538672 | 0.0E+00 |
|  |  | STDEV | 3.8E-12 | 0.0E+00 | 3.8E-12 | 0.0E+00 |
| g05 | 5126.496714 | BEST | 5126.496817 | -2.5E-14 | 5126.496714 | 8.9E-14 |
|  |  | AVERAGE | 5127.151053 | -2.5E-14 | 5126.496714 | -1.4E-14 |
|  |  | STDEV | 1.3E+00 | 0.0E+00 | 1.0E-12 | 3.6E-14 |
| g06 | -6961.813876 | BEST | -6961.813876 | 0.0E+00 | -6961.813876 | 0.0E+00 |
|  |  | AVERAGE | -6961.813876 | 0.0E+00 | -6961.813876 | 0.0E+00 |
|  |  | STDEV | 1.9E-12 | 0.0E+00 | 1.9E-12 | 0.0E+00 |
| g07 | 24.306209 | BEST | 24.330287 | 0.0E+00 | 24.306209 | 7.1E-15 |
|  |  | AVERAGE | 24.639188 | 0.0E+00 | 24.306209 | 5.2E-15 |
|  |  | STDEV | 2.4E-01 | 0.0E+00 | 1.1E-14 | 3.3E-15 |
| g08 | -0.095825 | BEST | -0.095825 | 0.0E+00 | -0.095825 | -1.7E-01 |
|  |  | AVERAGE | -0.095825 | 0.0E+00 | -0.095825 | -1.7E-01 |
|  |  | STDEV | 1.4E-17 | 0.0E+00 | 1.4E-17 | 6.3E-10 |
| g09 | 680.630057 | BEST | 680.630911 | 0.0E+00 | 680.630057 | 0.0E+00 |
|  |  | AVERAGE | 680.633029 | 0.0E+00 | 680.630057 | 8.2E-15 |
|  |  | STDEV | 1.6E-03 | 0.0E+00 | 7.6E-14 | 2.2E-14 |
| g10 | 7049.248021 | BEST | 7050.865659 | 0.0E+00 | 7049.248021 | 0.0E+00 |
|  |  | AVERAGE | 7093.127835 | 0.0E+00 | 7049.248024 | -3.2E-06 |
|  |  | STDEV | 3.0E+01 | 0.0E+00 | 1.0E-05 | 1.0E-05 |
| g11 | 0.749900 | BEST | 0.749900 | -8.4E-16 | 0.749900 | 4.4E-17 |
|  |  | AVERAGE | 0.749900 | -1.1E-16 | 0.749900 | 5.6E-18 |
|  |  | STDEV | 1.7E-08 | 2.6E-16 | 6.4E-17 | 6.4E-17 |
| g12 | -1.000000 | BEST | -1.000000 | 0.0E+00 | -1.000000 | -6.2E-02 |
|  |  | AVERAGE | -1.000000 | 0.0E+00 | -1.000000 | -6.2E-02 |
|  |  | STDEV | 0.0E+00 | 0.0E+00 | 0.0E+00 | 4.8E-15 |
| g13 | 0.053942 | BEST | 0.053942 | -6.2E-10 | 0.053942 | 6.2E-15 |
|  |  | AVERAGE | 0.053979 | -6.2E-11 | 0.053942 | 2.2E-15 |
|  |  | STDEV | 3.8E-05 | 2.0E-10 | 5.0E-17 | 2.2E-15 |
| g14 | -47.764888 | BEST | -47.723001 | -9.4E-06 | -47.764888 | 2.1E-16 |
|  |  | AVERAGE | -47.606122 | -9.9E-07 | -47.764888 | 1.9E-16 |
|  |  | STDEV | 1.5E-01 | 3.0E-06 | 1.1E-14 | 7.0E-17 |
| g15 | 961.715022 | BEST | 961.715023 | -3.9E-14 | 961.715022 | 3.3E-15 |
|  |  | AVERAGE | 961.715044 | -1.8E-14 | 961.715022 | 2.3E-15 |
|  |  | STDEV | 3.0E-05 | 1.6E-14 | 1.4E-13 | 2.4E-15 |
| g16 | -1.905155 | BEST | -1.905155 | 0.0E+00 | -1.905155 | 0.0E+00 |
|  |  | AVERAGE | -1.905155 | 0.0E+00 | -1.905155 | 0.0E+00 |
|  |  | STDEV | 4.7E-16 | 0.0E+00 | 4.7E-16 | 0.0E+00 |
| g17 | 8853.539675 | BEST | 8853.539675 | -1.1E-09 | 8853.539675 | -5.4E-14 |
|  |  | AVERAGE | 8853.539675 | -1.1E-10 | 8853.539675 | -2.9E-13 |
|  |  | STDEV | 1.3E-07 | 3.5E-10 | 1.3E-07 | 4.8E-13 |
| g18 | -0.866025 | BEST | -0.866014 | 0.0E+00 | -0.866025 | 6.8E-15 |
|  |  | AVERAGE | -0.858576 | 0.0E+00 | -0.866025 | 1.4E-15 |
|  |  | STDEV | 1.2E-02 | 0.0E+00 | 2.1E-15 | 2.5E-15 |
| g19 | 32.655593 | BEST | 34.879435 | 0.0E+00 | 32.655593 | 7.1E-15 |
|  |  | AVERAGE | 37.062544 | 0.0E+00 | 32.655593 | 4.8E-15 |
|  |  | STDEV | 1.4E+00 | 0.0E+00 | 4.1E-15 | 2.9E-15 |
| g20 | NaN | BEST | 0.080909 | 3.2E-03 | 0.177130 | 1.2E-01 |
|  |  | AVERAGE | 0.139964 | 1.8E-01 | 0.186264 | 3.3E-01 |
|  |  | STDEV | 3.8E-02 | 1.4E-01 | 3.4E-03 | 2.1E-01 |
| g21 | 193.724510 | BEST | 1113.283037 | – | NaN | NaN |
|  |  | AVERAGE | 1372.222391 | – | NaN | NaN |
|  |  | STDEV | 2.7E+02 | – | NaN | NaN |
| g22 | 236.430976 | BEST | 2144.075703 | 1.7E+00 | NaN | NaN |
|  |  | AVERAGE | 11776.390206 | – | NaN | NaN |
|  |  | STDEV | 9.4E+03 | – | NaN | NaN |
| g23 | -400.055100 | BEST | -2016.651110 | 1.7E+00 | -400.055100 | 3.3E-15 |
|  |  | AVERAGE | -966.309692 | 2.3E+00 | -400.055100 | 8.4E-15 |
|  |  | STDEV | 8.6E+02 | 2.6E-01 | 2.2E-13 | 1.2E-14 |
| g24 | -5.508013 | BEST | -5.508013 | 0.0E+00 | -5.508013 | 0.0E+00 |
|  |  | AVERAGE | -5.508013 | 0.0E+00 | -5.508013 | 0.0E+00 |
|  |  | STDEV | 9.4E-16 | 0.0E+00 | 9.4E-16 | 0.0E+00 |

Table 3: Results of the SQP supplied with the solutions of the GP-PSO. 'NaN' in problems g21 and g22 signify the lack of a suitable initial solution provided to the local search.

# 5 Conclusions

Combining an SQP local search to the GP-PSO algorithm as designed by Innocente et al.[4] has shown that it overcomes the inherent limitations in swarm dynamics of a stochastic swarm. It is also made apparent by the author of this document that an algorithm based solely on the features of a global optimizer is unlikely to achieve the best possible results across all problems. For this reason, a hybrid (Memetic algorithm) which implements individual learning, indicates considerable improvement of the solution quality, as well as the finding of the global solution in some cases, where swarm dynamics limit the statistical likelihood of it being found.

The GP-PSO-SQP has shown that it can compete with the leading optimizers, with its success rates across the problems of the test-suite. However, difficulties have been identified in its implementation, with its sensitivity to scaling and termination criteria. The computation expense of the algorithms by comparing problems successful on all three of the algorithms, indicates that the GP-PSO and the DMS-PSO to be at a similar level, and though PESO+ indicates a much higher FE count, its higher success on the range of problems tested overcomes any computational advantage that the GP-PSO may offer. The local search implementation however indicates that a great computational saving may be possible with suitable switching criteria and only a 2% computational expense on average of the total FEs by its implementation.